%% file: main.tex
\documentclass[10pt,twocolumn,letterpaper]{article}
\usepackage[pagenumbers]{iccv} 

\definecolor{iccvblue}{rgb}{0.21,0.49,0.74}
\usepackage[pagebackref,breaklinks,colorlinks,allcolors=iccvblue]{hyperref}

\usepackage{dsfont}
\usepackage{color, colortbl}
\usepackage{multirow}
\usepackage{arydshln}
\usepackage{pifont}
\usepackage{hyperref}
\usepackage{amsmath}
\usepackage{makecell}

\title{Robust Egocentric Visual Attention Prediction\\Through Language-guided Scene Context-aware Learning}

\author{Sungjune Park$^1$ \quad Hongda Mao$^2$ \quad Qingshuang Chen$^2$ \quad Yong Man Ro$^1$ \quad Yelin Kim$^2$ \\ $^1$Korea Advanced Institute of Science and Technology (KAIST) \quad $^2$Amazon \\
{\tt\small sungjune-p@kaist.ac.kr}
}

\begin{document}
\maketitle

\begingroup
\renewcommand\thefootnote{}
\footnotetext{This work was done during an internship at Amazon (Santa Cruz, CA, USA).}
\endgroup

\input{sec/0_abstract}
\input{sec/1_intro}
\input{sec/2_related}
\input{sec/3_method}
\input{sec/4_exp}

\input{sec/5_disc}
\input{sec/6_conc}

\bibliographystyle{ieeenat_fullname}
\bibliography{main}

\end{document}

%% file: sec/0_abstract.tex
\begin{abstract}
As the demand for analyzing egocentric videos grows, egocentric visual attention prediction, anticipating where a camera wearer will attend, has garnered increasing attention. However, it remains challenging due to the inherent complexity and ambiguity of dynamic egocentric scenes. Motivated by evidence that scene contextual information plays a crucial role in modulating human attention, in this paper, we present a language-guided scene context-aware learning framework for robust egocentric visual attention prediction. We first design a context perceiver which is guided to summarize the egocentric video based on a language-based scene description, generating context-aware video representations. We then introduce two training objectives that: 1) encourage the framework to focus on the target point-of-interest regions and 2) suppress distractions from irrelevant regions which are less likely to attract first-person attention. Extensive experiments on Ego4D and Aria Everyday Activities (AEA) datasets demonstrate the effectiveness of our approach, achieving state-of-the-art performance and enhanced robustness across diverse, dynamic egocentric scenarios.
\end{abstract}

%% file: sec/1_intro.tex
\section{Introduction}
\label{sec:intro}
With advancement of wearable camera technologies, egocentric vision has attracted growing interest, enabling comprehensive and accurate analysis of first-person view (FPV) videos~\cite{egoexo, egovlp, lego, egoexolearn, backpack}. A primary aspect of egocentric vision is to understand the events and interactions occurring around the camera wearer (i.e., the first person). In particular, it has emerged as a key challenge to predict where the camera wearer is likely to focus--referred to as egocentric visual attention prediction. It aims to anticipate point-of-interest (PoI) regions\footnotemark, which are often closely associated with the first-person's interactions with surrounding objects and people~\cite{eye1, eye2, eye3}.

\begin{figure}[t]
  \centering
  \includegraphics[width=0.999\linewidth]{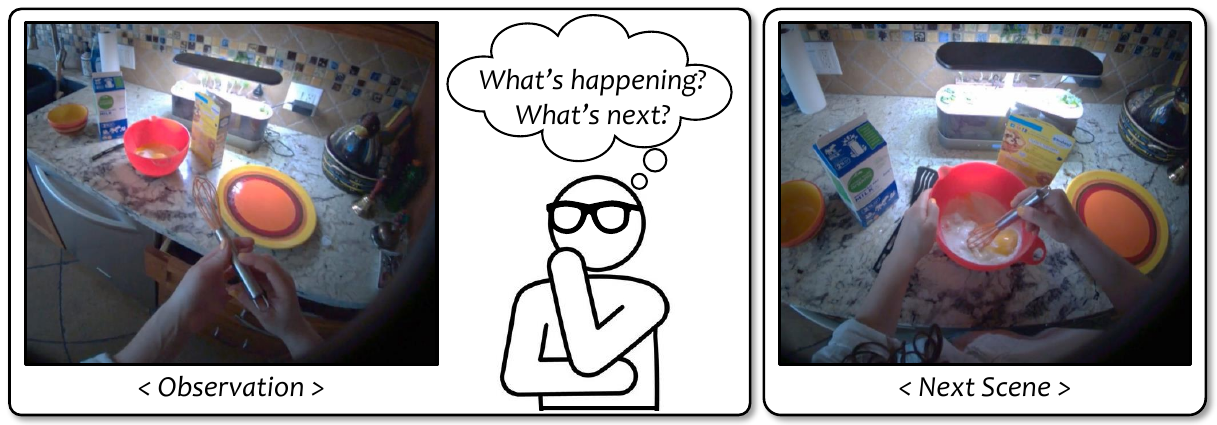}
   \caption{An example showing how contextual cues help predict the point-of-interest region. When humans observe the given scene (\textit{left}), humans can understand the scene context--a red bowl with an egg mixture and a whisk in hand. Therefore, humans easily infer that the red bowl will likely become the focus of first-person attention.}
   \label{fig1}
\end{figure}

\footnotetext{The PoI regions refer to semantically meaningful and important regions that the camera wearer is likely to attend.}

However, it remains difficult to accurately predict these attention regions, because the egocentric video scene itself is inherently complicated and ambiguous due to its dynamic environments and varying contextual cues~\cite{csts}. To address the problem, various approaches have been proposed to improve the robustness in egocentric visual attention prediction~\cite{attntrans, gaze2, gaze3, gaze6}\footnote{These works include visual attention estimation and anticipation (prediction) research. The attention anticipation extends the estimation to further predict future point-of-interest (PoI), not limited to the current frame.}. For instance, Li et al.~\cite{gaze1} leverage auxiliary cues such as head motion and hand positions to localize the first-person's attention. GazeMLE~\cite{gazemle} explores the relationship between first-person attention and action by utilizing both RGB and optical flow inputs to jointly predict visual attention and actions. Similarly, Global-Local Correlation (GLC)~\cite{glc} considers interactions between global video frames and local regions. Lai et al.~\cite{csts} further incorporate audio information, because humans often attend toward sound sources.

Different from the existing methods, our approach is inspired by a human visual cognitive theory: scene context serves as a critical stimulus guiding humans' attention and eye movement~\cite{context1, context2, context3}. Based on this insight, we argue that it would be beneficial to understand the overall scene context for anticipating the PoI regions effectively. For instance, Figure~\ref{fig1} illustrates how contextual information helps predict where the first person is likely to attend to. While observing various objects--such as an empty plate and a red bowl with an egg mixture, we can also recognize that the camera wearer is cooking, holding a whisk, and approaching the cooking table. Given such contextual cues, it is natural to anticipate that the first-person visual attention will be directed toward the red bowl in order to whip the egg mixture. However, despite the importance of such contextual cues, existing methods often fail to explicitly consider such scene context, and instead incorporate auxiliary modality--such as motion or audio data--to implicitly capture missing contextual information.

To bridge this gap, in this paper, we propose a novel language-guided, scene context-aware learning framework for robust egocentric visual attention prediction. Our framework is designed to comprehend the overall scene context and enhance focus on the target PoI regions while minimizing distractions from irrelevant areas. Specifically, we introduce a context perceiver composed of a context summary extractor and a context summary guider. The context summary extractor generates video summary representations which capture the overall context of the input egocentric video, guided by a language-based scene description. The context summary guider then refines the video features leveraging these context representations, making them context-aware. After guiding to understand the overall scene context, we further introduce two training objectives: a negative region loss and a region suppression loss, which help the framework focus on the target attention regions. The negative region loss encourages the framework to predict precise attention heat maps by clearly separating target PoI regions from attention-irrelevant areas. Meanwhile, the region suppression loss aims to minimize distraction by reducing activations on irrelevant regions and ensuring focus on the target regions. We validate the proposed framework through extensive experiments on two public egocentric video datasets, Ego4D \cite{ego4d} and Aria Everyday Activities (AEA) \cite{aria}, achieving state-of-the-art performance. Our contributions are summarized as follows:
\begin{itemize}
    \item We introduce a novel language-guided, scene context-aware learning framework understanding the overall scene contextual cues for robust egocentric visual attention prediction.
    \item We propose a context perceiver along with two training objectives which jointly enhance scene context understanding and visual attention localization.
    \item The comprehensive experiments on Ego4D and AEA datasets demonstrate the effectiveness and robustness of our method, achieving state-of-the-art performance.
\end{itemize}

%% file: sec/2_related.tex
\section{Related Work}
\label{sec:related}

\subsection{Egocentric Vision}
The growing demand of analyzing first-person video (FPV)has led to the rapid development of egocentric vision. Recent large-scale egocentric video benchmarks have further accelerated its advancement~\cite{egovis1, egovis2, egovis3, epic, egoexolearn}. Lin et al.~\cite{egovlp} pioneered EgoVLP, a pretraining method that jointly align egocentric video and language using large-scale Ego4D data~\cite{ego4d}. Li et al.~\cite{egoexo} investigated egocentric properties observed in third-person videos and proposed to distill such knowledge from exocentric videos for the effective egocentric pretraining. Several studies have also explored modeling scene context to solve various downstream tasks in egocentric vision~\cite{egovis4, egovis6, egovis7, egovis9, egovis10}. For instance, Kazakos et al.~\cite{egovis8} incorporated audio signals to enhance visual action representations in temporal contexts. Thakur et al.~\cite{egovis5} introduced a context-aware modeling for short-term object interaction by detecting objects, analyzing their motions, and capturing the relationship between objects and global video features. Lai et al.~\cite{lego} raised a novel egocentric action frame generation problem and conditioned a diffusion model on the user's context to synthesize egocentric action frames. Nagarajan et al.~\cite{egovis10} aimed to learn environment-aware video representations by storing the semantics of local surroundings of the camera wearer, including objects outside the visible view. As these works demonstrate, contextual information is considered as important and leveraged for each purpose and framework. However, despite the clear effectiveness of scene context in egocentric vision, it has not been explored yet for egocentric visual attention prediction.

\subsection{Egocentric Visual Attention Prediction}
Since the point-of-interest (PoI) regions in egocentric videos provide critical information about the camera wearer's intent and interaction, egocentric visual attention prediction has received increasing attention~\cite{eye1, eye2, eye3}. This task includes two problems, attention estimation and anticipation. The attention estimation predicts the PoI regions in the current frame~\cite{attntrans, gaze4, gazemle, gaze5, glc}, and the attention anticipation further predicts the visual attention regions in the future frames~\cite{dfg, dfg+, csts}. For example, Lai et al.~\cite{glc} proposed extracting a global video token via simple pooling across video frames, which is them integrated with local frame tokens. In addition, several methods have exploited auxiliary modality cues to understand the complicated egocentric scene information. Huang et al.~\cite{attntrans} employed optical flow alongside RGB video data, building two separate pathways to encode motion and visual features. Li et al.~\cite{gazemle} jointly used RGB and flow input frames together to obtain motion-aware visual features, training the model to predict attention maps and recognize actions simultaneously. Thakur et al.~\cite{gaze5} adopted inertial measurement unit (IMU) data to complement the insufficient context cues in the egocentric scene. More recently, Lai et al.~\cite{csts} combined audio and video features to identify sound sources and improve attention prediction.

However, these approaches overlook the importance of scene context understanding, not explicitly capturing contextual cues. Additionally, previous methods lack of considering how to focus on the target attention region without being distracted from other irrelevant regions. In this paper, we present a novel language-guided, scene context-aware learning framework for robust egocentric visual attention prediction, which is designed to: 1) understand the overall scene context and 2) accurately focus on the target PoI regions while reducing distractions from the irrelevant areas.

%% file: sec/3_method.tex
\section{Language-guided Scene Context-aware Learning}
\label{sec:method}
In this section, we describe our language-guided, scene context-aware learning framework. We begin by preparing scene summary descriptions and their embeddings, which summarize egocentric videos and serve as semantic guidance. The following subsections elaborate how to acquire such scene summary descriptions and embeddings. We then explain the context perceiver, which aims to obtain contextualized video features. Finally, we introduce the proposed two training objectives--a negative region loss and a region suppression loss--designed to encourage the framework to concentrate on the target PoI region while suppressing distraction impacts from other regions irrelevant with the first-person attention.

\subsection{Scene Summary Description Preparation}
\label{summary}
\begin{figure}[t]
  \centering
  \includegraphics[width=0.999\linewidth]{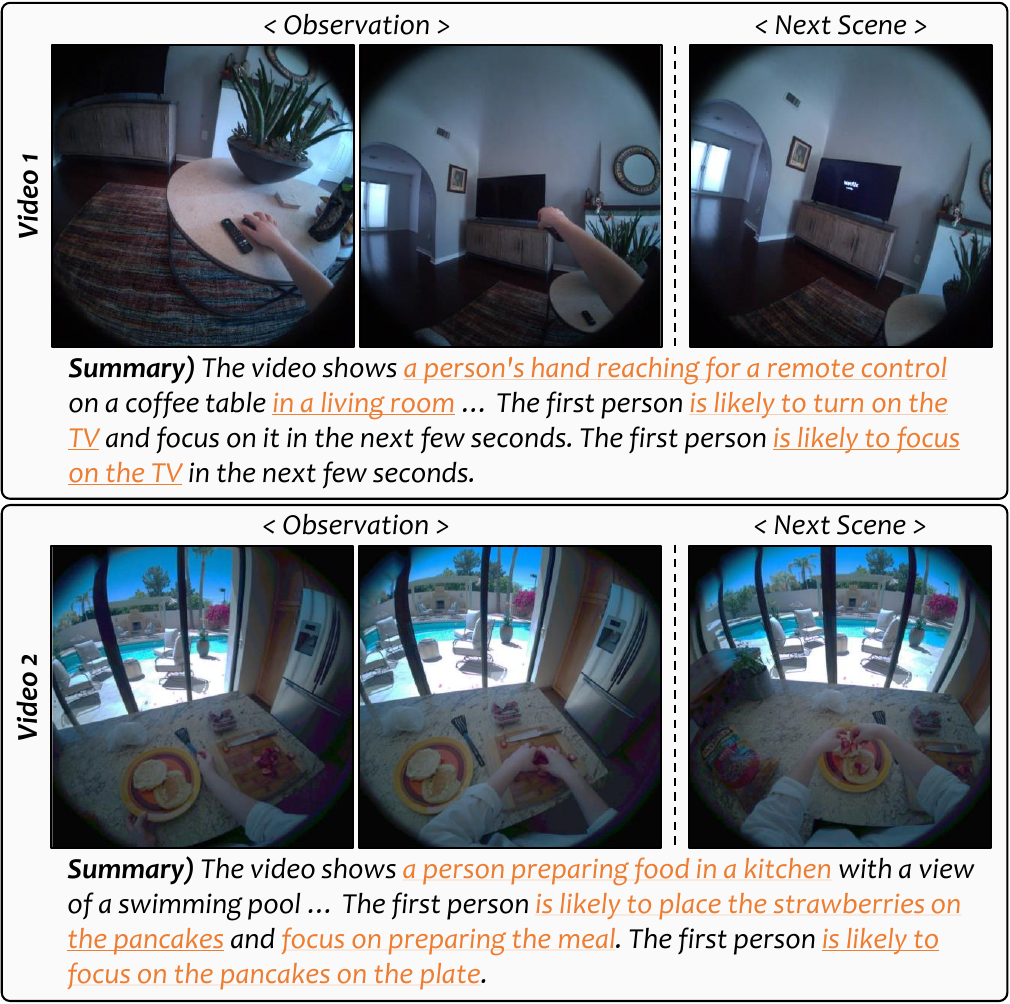}
   \caption{The examples of scene summary descriptions, which include location, action, and object information (e.g., living room, reaching for a remote control, and TV) related with the first person.}
   \label{fig2}
\end{figure}

Before training the proposed context-aware egocentric visual attention prediction framework, we prepare video scene descriptions which semantically summarize egocentric videos. To this end, we adopt VideoChat2~\cite{videochat2}, a video-based dialogue model capable of understanding egocentric videos. We prompt the model with a chain of instructions for each input video as follows: \textbf{P1:} {\tt Please describe the details of the video.}, \textbf{P2:} {\tt Which action is the first person likely to perform and focus on in the next few seconds?}, \textbf{P3:} {\tt Which object is the first person likely to focus on in the next few seconds?}. These prompts are designed to elicit descriptions that reflect the scene context relevant to the first person. Figure~\ref{fig2} provides examples of generated descriptions, which effectively summarize where the first person is, what the first person is doing, and what the first person is likely to focus on. For instance, the second example describes a kitchen scene where the person is preparing a meal with strawberries and will likely focus on the pancake. After obtaining the scene descriptions, we extract their embeddings using the pretrained sentence embedding model NV-Embed-v2~\cite{nv-embed-v2}. These embeddings are used to guide the context perceiver, as described in the following subsection. This entire preparation step is performed prior to training the egocentric visual attention prediction framework.

\begin{figure*}[t]
  \centering
  \includegraphics[width=0.9\linewidth]{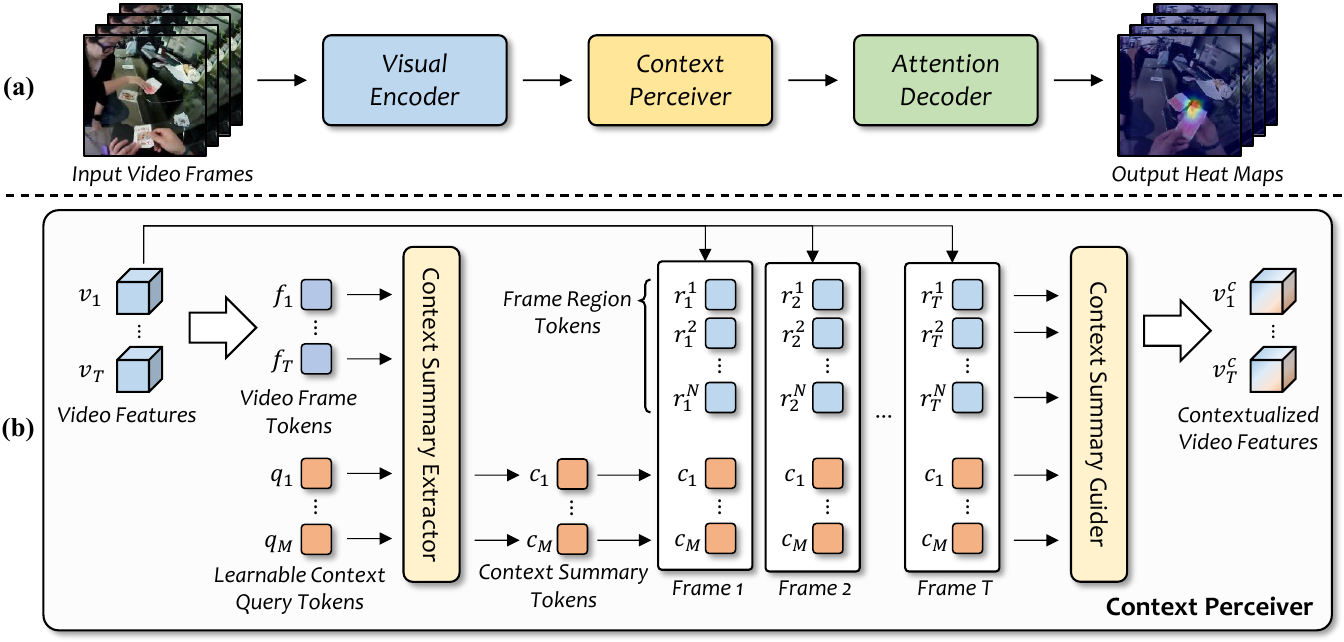}
   \caption{(a) The overall architecture of egocentric visual attention anticipation framework along with the proposed context perceiver. (b) The details of context perceiver consisting of context summary extractor and guider. The context summary extractor is designed to obtain context summary representations by looking through whole video frames, and the context summary guider helps frame region tokens refer to the scene context summary tokens for each frame respectively. During the training, context summary tokens are guided to represent the scene summary description embedding which is paired with the input video.}
   \label{fig3}
\end{figure*}

\subsection{Context Perceiver}
\label{cp}
Figure~\ref{fig3}(a) illustrates the overall architecture of the egocentric visual attention prediction framework. Given an input egocentric video frames, the visual encoder first extracts frame-wise video features, which are then passed through the context perceiver. These vanilla features become the contextualized (i.e., context-aware) video features which capture the overall scene context. The attention decoder subsequently predicts the attention heat maps indicating where the first person will focus. 

Figure~\ref{fig3}(b) shows the details of the context perceiver, composed of the context summary extractor and the context summary guider. The extractor aims to generate the context summary tokens which properly represent the given video. The guider helps video frame tokens refer to these context summary tokens to obtain the contextualized video features. The context perceiver takes the video features $\boldsymbol{V}=\{\boldsymbol{v_t}\}^T_{t=1} \in \mathbb{R}^{T \times C \times H \times W}$ from the encoder, where $T$ is the video time dimension (i.e., frame), and $H$, $W$, and $C$ are the height, width, and channel dimensions, respectively. A 3D convolution is applied to obtain the video frame tokens $\boldsymbol{F}=\{\boldsymbol{f_t}\}^{T}_{t=1} \in \mathbb{R}^{T \times C}$. Along with $\boldsymbol{F}$, we also place learnable context query tokens $\boldsymbol{Q}=\{\boldsymbol{q_m}\}^M_{m=1} \in \mathbb{R}^{M \times C}$, where $M$ is the number of the queries, in order to read the video content of $\boldsymbol{F}$. Both frame tokens $\boldsymbol{F}$ and query tokens $\boldsymbol{Q}$ are then fed into the context summary extractor, that is a self-attention layer. It enables $\boldsymbol{Q}$ to refer to each content of $\boldsymbol{F}$. The resulting output tokens are the set of context summary tokens $\boldsymbol{C}=\{\boldsymbol{c_m}\}^M_{m=1} \in \mathbb{R}^{M \times C}$. To guide $\boldsymbol{C}$ to include the proper scene context information with the given video, we utilize the corresponding scene summary description embedding $\boldsymbol{d}$. Then we find the most similar summary token $\boldsymbol{c_s}$ to $\boldsymbol{d}$ among the $M$ context summary tokens $\boldsymbol{C}$. Using this most relevant summary token $\boldsymbol{c_s}$, we design a context encoding loss to effectively incorporate the scene context information of $\boldsymbol{d}$ as follows:
\begin{equation}
    \mathcal{L}_{context} = -\text{log}(\text{sim}(\boldsymbol{c_s}, \boldsymbol{d})),
\label{eq1}
\end{equation}
\noindent
where $\text{sim}()$ denotes cosine similarity function, and $s$ is the index with the highest similarity to the description $\boldsymbol{d}$.

After that, the context guider refines each frame's region tokens with the obtained scene context. We feed $\boldsymbol{C}$ into the context summary guider along with the frame region tokens. The region tokens of the $k$-th frame is denoted as $\boldsymbol{R_k}=\{\boldsymbol{r^p_k}\}^N_{p=1} \in \mathbb{R}^{N \times C}$, where $N$ is the number of region tokens ($=H \times W$). The guider is based on the in-frame self-attention mechanism, not referring to the region tokens from the other frames. For instance, the region tokens from the $k$-th video frame $\boldsymbol{R_k}$ do not attend to the region tokens from all the other video frames. Instead, $\boldsymbol{R_k}$ attends to its own tokens along with the context summary tokens $\boldsymbol{C}$. After that, we obtain contextualized video features $\boldsymbol{V^C}=\{\boldsymbol{v^c_t}\}^T_{t=1} \in \mathbb{R}^{T \times C \times H \times W}$ which perceives the overall scene context across entire video frames.

\subsection{Minimizing Distraction Impact}
\label{loss}
\begin{figure}[t]
  \centering
  \includegraphics[width=0.999\linewidth]{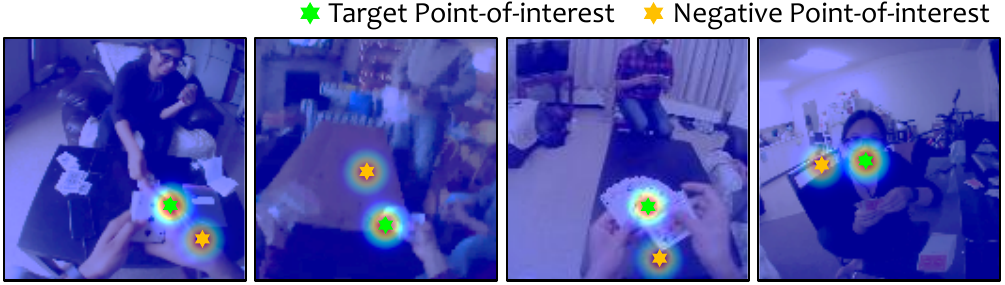}
   \caption{The examples of the negative points located around the target point.}
   \label{fig4}
\end{figure}

To encourage the framework to concentrate on the target PoI regions and minimize distractions from the other attention-irrelevant areas, we design two training objectives: \textbf{a negative region loss} and \textbf{a region suppression loss}. The negative region loss is designed to ensure the predicted attention heat map is distinguishable from pseudo-negative samples in the contrastive manner. The pseudo-negative samples include attention-irrelevant regions spatially located near the target point. We sample negative points around the target attention point with random radius and angle. In Figure~\ref{fig4}, the green and yellow dots denote the examples of target and pseudo negative points, respectively. Then each point--both target and negative points--is then assumed to be the center of a 2D Gaussian distribution along the vertical and horizontal axes, forming heat map distributions. We denote the target attention distribution and the negative attention distribution at the $k$-th video frame as $\boldsymbol{A_k} \in \mathbb{R}^{HW}$ and $\boldsymbol{A^n_k} \in \mathbb{R}^{HW}$, respectively. Additionally, there also exists the predicted attention heat map distribution $\boldsymbol{\hat{A}_k} \in \mathbb{R}^{HW}$. By using these three distributions, we calculate the positive similarity $s^p_k$ and the negative similarity $s^n_k$ for the $k$-th video frame as follows:
\begin{equation}
    s^p_k = \text{sim}(\boldsymbol{\hat{A}_k}, \boldsymbol{A_k}), \quad s^n_k = \text{sim}(\boldsymbol{\hat{A}_k}, \boldsymbol{A^n_k}).
\label{eq2}
\end{equation}

\noindent
Based on the similarities, we formulate the negative region loss for every video frame as follows:
\begin{equation}
    \mathcal{L}_{neg} = -\frac{1}{T} \sum\limits_{t=1}^T \text{log}(\frac{\text{exp}(s^p_t / \tau)}{\text{exp}(s^p_t / \tau) + \sum_{i=1}^{N'} \text{exp}(s^n_{t, i} / \tau)}),
\label{eq3}
\end{equation}
\noindent
where $T$ is the number for video frames, $\tau$ is the temperature for contrastive learning, and $N'$ is the number of negative samples per frame. By doing so, we guide the predicted attention heat map distribution to be similar with the target distribution, while making it distinct from the negative distributions. Thus, the framework is trained to concentrate on the target areas, while excluding the other irrelevant regions.

\input{tables/table1}

To further eliminate irrelevant activations, we introduce the region suppression loss. Ideally, the predicted attention heat map distribution should be highly activated on the target regions only, while having zero probabilities on the other areas. To separate target and attention-irrelevant regions, we generate a binary region mask. Given the target distribution $\boldsymbol{A_k}$, the region mask for the $k$-th video frame can be $\text{\textbf{M}}_{\boldsymbol{k}}=\mathds{1}[\boldsymbol{A_k} \geqq \delta] \in \mathbb{R}^{HW}$. Here, $\delta$ is the threshold to mask out the other attention-irrelevant areas. For every video frame, the region suppression loss $\mathcal{L}_{supp}$ is formulated as follows:
\begin{multline}
    \mathcal{L}_{supp} = -\frac{1}{T} \sum\limits_{t=1}^T \sum\limits_{j=1}^{HW} \text{M}^j_t \cdot A^j_t \, \text{log} (\hat{A}^j_k) \\+ (1 - \text{M}^j_t) \cdot (1 - A^j_t) \, \text{log}(1 - \hat{A}^j_k).
\label{eq4}
\end{multline}
\noindent
Due to $\mathcal{L}_{supp}$, the probabilities on the attention regions will be higher, while the other residual areas become lower. In other words, it encourages high activation on PoI regions and suppresses other regions to be predicted. Finally, we incorporate all the proposed training losses with the KL divergence loss~\cite{glc, csts} which is generally adopted in egocentric visual attention prediction.

%% file: tables/table1.tex
\definecolor{Blue}{rgb}{0.9, 0.95, 0.97}

\begin{table*}[t]
    \centering
    \renewcommand{\tabcolsep}{5.0mm}
    \resizebox{0.99\linewidth}{!}
    {\small
        \renewcommand{\arraystretch}{1.3}
        \begin{tabular}{lcccccc}
            \toprule[1.2pt]
            \multicolumn{1}{c}{\multirow{2}{*}{\bf Methods}} & \multicolumn{3}{c}{\bf Ego4D} & \multicolumn{3}{c}{\bf Aria Everyday Activities (AEA)} \\ \cmidrule(lr){2-4} \cmidrule(lr){5-7}
            \multicolumn{1}{c}{} & \multicolumn{1}{c}{\bf F1 Score} & \multicolumn{1}{c}{\bf Recall} & \multicolumn{1}{c}{\bf Precision} & \multicolumn{1}{c}{\bf F1 Score} & \multicolumn{1}{c}{\bf Recall} & \multicolumn{1}{c}{\bf Precision} \\
            \midrule
            GazeMLE (PAMI'21) $^{flow}$ & 36.3 & 52.5 & 27.8 & 56.8 & 64.1 & 51.0 \\
            AttnTrans (ECCV'18) $^{flow}$ & 37.0 & 55.0 & 27.9 & 57.4 & 65.5 & 51.0 \\
            CSTS (ECCV'24) $^{audio}$ & 39.7 & 53.3 & 31.6 & 59.9 & 66.8 & 54.3 \\ \hdashline
            I3D-R50 (ICCV'19) & 36.9 & 52.1 & 28.6 & 57.4 & 63.6 & 52.2 \\
            DFG (CVPR'17) & 37.2 & 53.2 & 28.6 & 57.4 & 63.6 & 52.3 \\
            MViT (ICCV'21) & 37.2 & 54.1 & 28.3 & 57.5 & 62.4 & 53.3 \\
            DFG+ (PAMI'18) & 37.3 & 52.3 & 29.0 & 57.6 & 65.5 & 51.3 \\
            GLC (BMVC'22) & 37.8 & 52.9 & 29.4 & 58.3 & 65.4 & 52.6 \\
            \midrule
            \rowcolor{Blue}
            \bf Ours & \textbf{40.1} & \textbf{54.1} & \textbf{31.9} & \textbf{60.3} & \textbf{67.2} & \textbf{54.7} \\
            \bottomrule[1.2pt]
        \end{tabular}
    }
    \caption{Performance comparison with state-of-the-art methods on Ego4D and AEA testing sets, respectively. The superscripts $flow$ and $audio$ denote the methods which utilize additional sensory modality input during the inference time, while the other methods employ the video frame only.}
    \label{tab1}
\end{table*}

%% file: sec/4_exp.tex
\section{Experiment}
\label{sec:exp}
\subsection{Experimental Setup}
\subsubsection{Evaluation Benchmarks}
We evaluate our framework on two public egocentric video benchmarks: Ego4D~\cite{ego4d} and Aria Everyday Activities (AEA)~\cite{aria}.

\textbf{Ego4D} is one of the large egocentric video datasets. We use subset data which includes eye-tracking annotations, containing 27 egocentric videos collected in social interactive settings, such as playing card games with partners. Following Lai et al.~\cite{glc}, we split the video data into 15,310 training video clip segments from 20 videos and 5,202 testing video clip segments from the remaining 7 videos. Each video clip segment is 5 seconds long, and these segments are not overlapped each other.

\textbf{AEA} includes 143 egocentric videos that record various indoor daily activities, such as social interaction, cooking, and cleaning. The total duration of entire video data is approximately 7 hours with 20 fps having a 1408$\times$1408 resolution. Following the data split used by Lai et al.~\cite{csts}, we use 107 videos for the training and 36 videos for the testing. As a result, each set includes 10,456 and 2,901 video clips, respectively, where each video clip is also 5 seconds long.

For the evaluation, we follow evaluation protocols in prior works~\cite{gazemle, glc, csts}, considering the visual attention anticipation task as a binary classification problem on each 2D video frame. We use the F1 score as the primary evaluation metric, and report corresponding recall and precision together. We do not adopt AUC score as the evaluation metric due to the performance saturation issue caused by the long-tailed distribution characteristics of attention points on 2D frame~\cite{glc, csts}.

\subsubsection{Implementation Details}
For the implementation, according to Lai et al.~\cite{csts}, we divide each video clip into a 3 second-long segment as input (i.e., the given past frames) and a 2 second-long segment for future prediction. We uniformly sample 8 frames from the input segment, and the attention decoder generates the visual attention heat maps for 8 uniformly sampled future frames corresponding to the 2 second-long future segment. Every input video is resized to 256$\times$256 resolution for both Ego4D and AEA datasets. Our framework is implemented based on PySlowFast~\cite{slowfast}. We use MViT self-attention blocks~\cite{mvit} as the visual encoder and the transformer-based attention decoder. The context pereceiver is composed of one self-attention block each for the summary extractor and the guider. To extract video frame tokens, we use 3D convolution with $1 \times 3 \times 3$ kernel and a dilation spacing of $2$. We use $M=32$ learnable context query tokens. The framework is trained for 15 epochs using AdamW optimizer with a base learning rate of $1e^{-4}$, momentum of $0.9$, and weight decay of $5e^{-2}$.

\begin{figure*}[!t]
  \centering
  \includegraphics[width=0.999\linewidth]{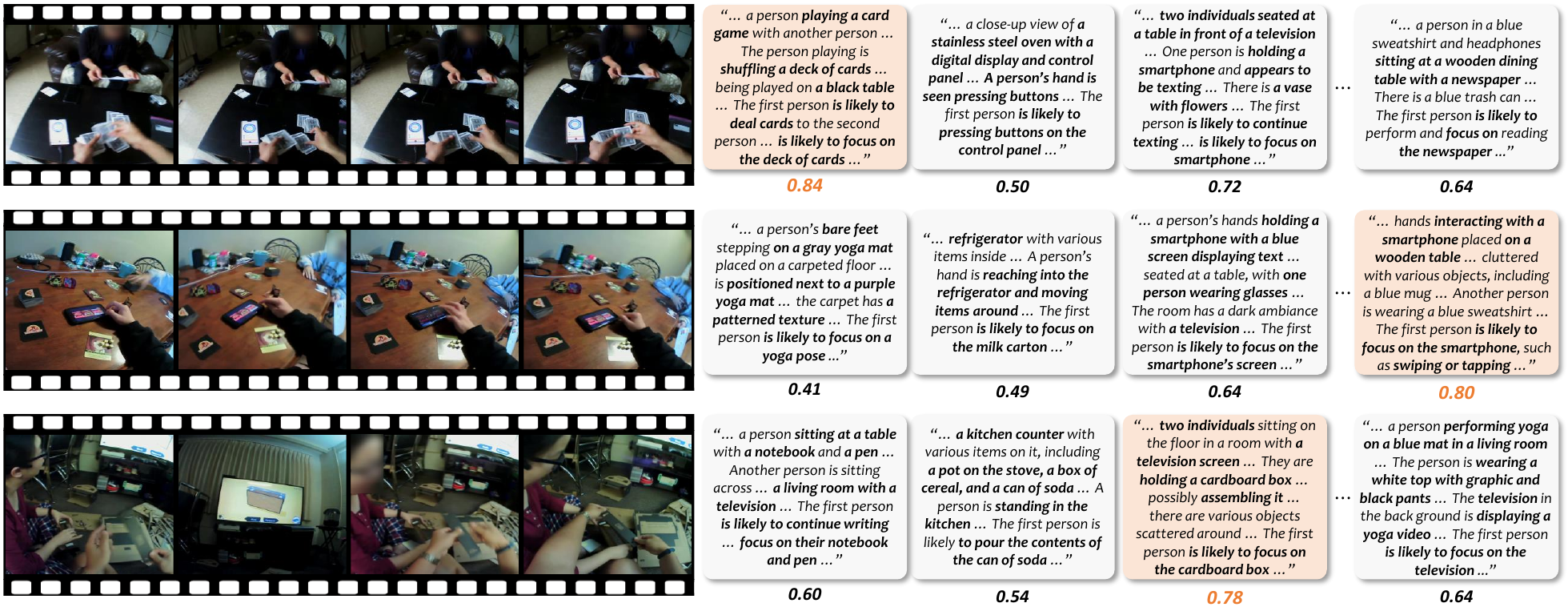}
    \caption{The examples show whether the context perceiver can properly capture scene context representation. When the input video is observed, the context perceiver outputs context summary tokens. Then we compare the similarity between the tokens and summary description embeddings. In the first row, the context summary token is well matched with the corresponding scene summary description showing the highest similarity 0.84.}
   \label{fig5}
\end{figure*}

\subsection{Experimental Result}
\subsubsection{Comparison with State-of-the-arts}
\label{comp}
To validate the effectiveness of our approach, we compare its performance with existing state-of-the-art methods on both Ego4D and AEA. These include GazeMLE~\cite{gazemle}, AttnTrans~\cite{attntrans}, CSTS~\cite{csts}, I3D-R50~\cite{i3d}, DFG~\cite{dfg}, MViT~\cite{mvit}, DFT+~\cite{dfg+}, and GLC~\cite{glc}. The competitive prior performances are referred to the state-of-the-art method~\cite{csts}. As shown in Table~\ref{tab1}, our method achieves remarkable performances on both datasets: 40.1 F1, 54.1 Recall, and 31.9 Precision on Ego4D, and 60.3 F1, 67.2 Recall, and 54.7 Precision on AEA. The proposed framework clearly outperforms the prior methods which utilize egocentric visual information only. Notably, even compared to the methods which employ additional information at inference time (e.g., optical flow and audio), our framework--guided by language-based scene context only during training--achieves superior performance on both datasets.

\input{tables/table2}

\subsubsection{Performance on Unseen Videos}
\label{zero}
To validate generalizability, we conduct a zero-shot experiment. Following CSTS~\cite{csts}, we train the proposed framework using the Ego4D training data and evaluate it on the unseen AEA test set. Table~\ref{tab2} presents performance comparisons under the same zero-shot experimental setting, where the methods are ordered by F1 score which is the primary evaluation metric. As shown in the table, the proposed framework achieves noticeable performance of 53.7 F1 score. According to this experimental result, our framework, with scene-context understanding, demonstrates superior generalization to unseen egocentric scenarios than prior works.

\input{tables/table3}

\subsubsection{Context Summary Token Analysis}
In this subsection, we examine whether the context summary extractor appropriately produces the meaningful scene context summary tokens $\boldsymbol{C} \in \{\boldsymbol{c_m}\}^{M}_{m=1}$. To analyze it, we compare the similarity between the summary tokens and the description embeddings from various egocentric videos. These embeddings include descriptions of the given input video and the other egocentric videos. Specifically, we compute the cosine similarity between each context summary token and the description embedding. Figure~\ref{fig5} shows three video examples. In the first example, the correct description yields the highest similarity (0.84), while others remain low. This analysis corroborates that the context summary extractor properly captures the semantic context of the input egocentric video, thereby validating the role of the context perceiver in generating context-aware video features.

\subsubsection{Ablation Study}
We conduct ablation study to verify the effectiveness of each proposed component: the negative region loss, region suppression loss, and context perceiver. We evaluate F1 score on Ego4D and AEA by selectively adopting each component. As shown in Table~\ref{tab3}, each component improves performance over the baseline. The baseline is the re-implementation of MViT~\cite{mvit}, and it does not adopt any of the proposed components. On Ego4D dataset, the negative region loss, suppression loss, and the context perceiver contribute F1 improvements of 2.0, 2.0, and 2.1, respectively. When all components are used together, the entire framework achieves an overall F1 improvement of 2.7. Additionally, on AEA dataset, each component brings performance improvements, obtaining 1.1, 1.7, and 1.7  in F1 score, respectively. With three components, it achieves 2.6 F1 score improvement from the baseline.

\begin{figure*}[t]
  \centering
  \includegraphics[width=0.999\linewidth]{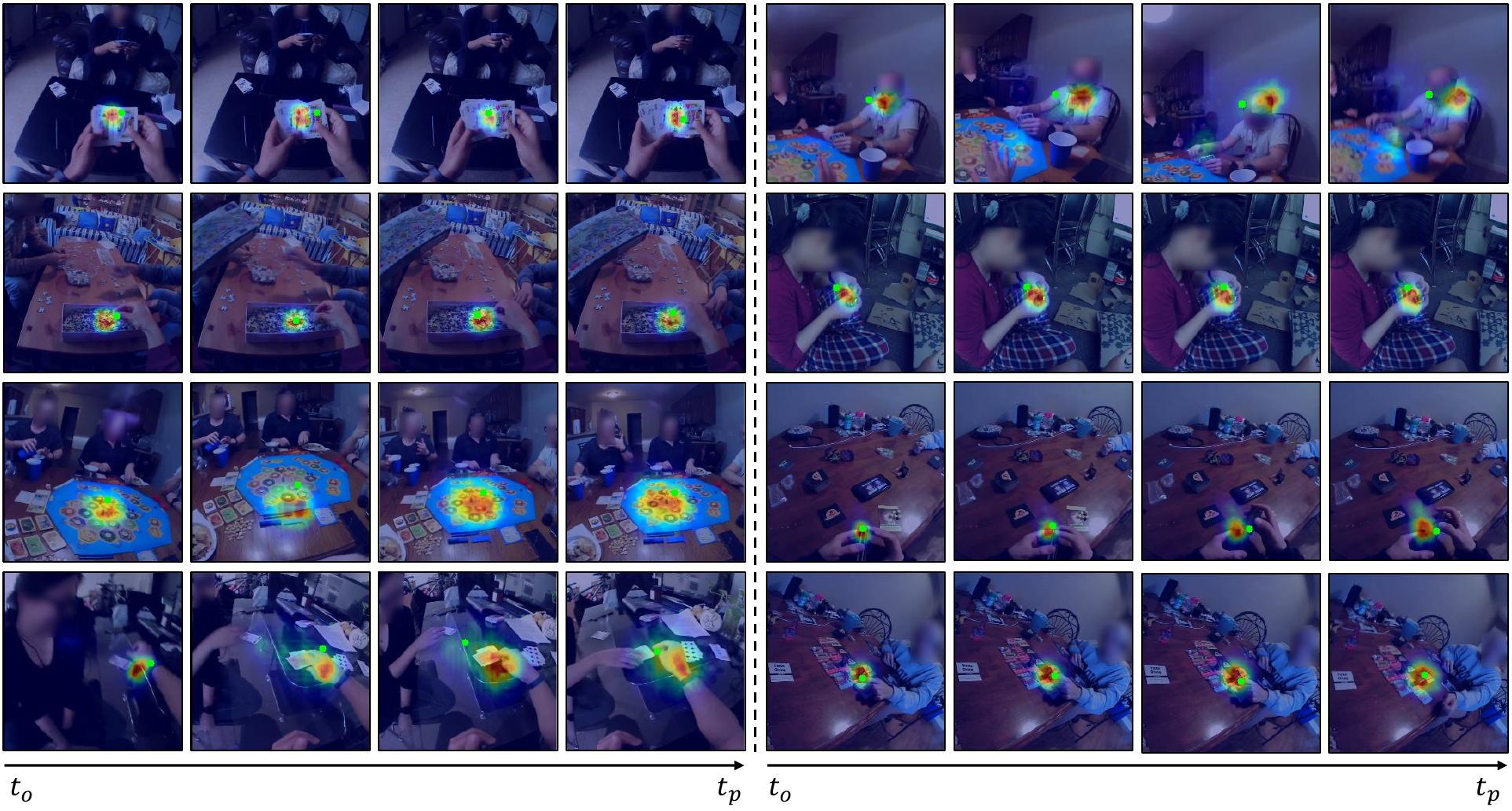}
    \caption{The qualitative visualization results on Ego4D dataset. The green dots indicate the target ground-truth points, and the heat maps represent the predicted attention maps. Each output displays four sampled video frames in a time series.}
   \label{fig6}
\end{figure*}

\begin{figure}[t]
    \centering
    \includegraphics[width=0.9\linewidth]{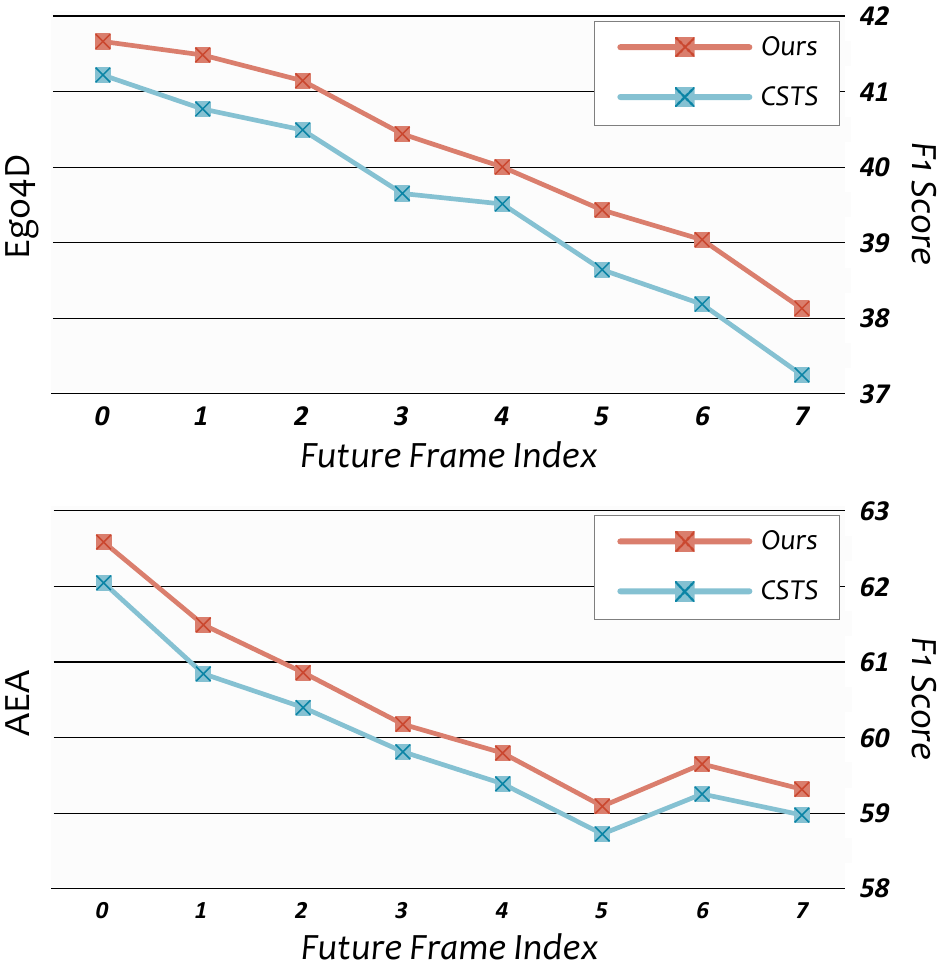}
    \caption{The performance analysis on each future frame and comparison with CSTS~\cite{csts} on 8 future frames.}
    \label{fig7}
\end{figure}

\subsubsection{Visualization Results}
Figure~\ref{fig6} presents 8 visualization examples from Ego4D dataset~\cite{ego4d}. Given a 3 second-long past video segment as input, our framework predicts the visual attention over the future 2 seconds. The figure shows the predicted attention heat maps for future video frames. In the figure, $t_0$ and $t_p$ represent the current and future prediction timestamps, respectively, such that $t_p - t_0 = 2s$. The green dots denote the ground-truth PoIs, and each row corresponds to a different example. As illustrated in the figure, our framework effectively predicts future attention in various situations focusing on (i) focusing on where the first person is doing by their hands, (ii) attending to the partner who the first person is engaging with, and (iii) understanding what their partner is interacting with.

\subsubsection{Performance Analysis Per Future Frame}
Following the experimental setup in CSTS~\cite{csts}, our framework anticipates first-person PoI regions across 8 future frames (corresponding to the next 2 seconds) using the given input 3 second-long video. We analyze the performance for each future frame, and Figure~\ref{fig7} plots F1 score graphs on both Ego4D and AEA, respectively. For the comparison, the figure includes performance of CSTS~\cite{csts}, which uses audio sensor data at inference time. Based on the analysis, our method, utilizing language guidance for scene context-aware learning, shows robust performance on future frames.

%% file: tables/table2.tex
\definecolor{Blue}{rgb}{0.9, 0.95, 0.97}

\begin{table}[!t]
    \centering
    \renewcommand{\tabcolsep}{2.0mm}
    \resizebox{0.99\linewidth}{!}
    {\small
        \renewcommand{\arraystretch}{1.3}
        \begin{tabular}{lccc}
            \toprule[1.2pt]
            \bf Methods & \bf F1 Score & \bf Recall & \bf Precision \\
            \midrule
            DFG (CVPR'17) & 39.3 & 80.4 & 26.0 \\
            I3D-R50 (ICCV'19) & 41.5 & 77.2 & 28.4 \\
            AttnTrans (ECCV'18) $^{flow}$ & 43.1 & 57.5 & 34.5 \\
            DFG+ (PAMI'18) & 43.1 & 76.4 & 30.0 \\
            GazeMLE (PAMI'21) $^{flow}$ & 44.0 & 59.0 & 35.0 \\
            MViT (ICCV'21) & 44.1 & 59.7 & 35.0 \\
            GLC (BMVC'22) & 46.9 & 72.8 & 34.6 \\
            CSTS (ECCV'24) $^{audio}$ & 50.8 & 62.2 & 42.9 \\
            \midrule
            \rowcolor{Blue}
            \bf Ours & \bf 53.7 & \bf 65.0 & \bf 45.7 \\
            \bottomrule[1.2pt]
        \end{tabular}
    }
    \caption{Performance comparison under zero-shot unseen video experimental setting.}
    \label{tab2}
\end{table}

%% file: tables/table3.tex
\definecolor{Blue}{rgb}{0.9, 0.95, 0.97}
\newcommand{\cmark}{\ding{51}}
\newcommand{\xmark}{\ding{55}}

\begin{table}[t]
    \centering
    \renewcommand{\tabcolsep}{1.5mm}
    \resizebox{0.99\linewidth}{!}
    {\small
        \renewcommand{\arraystretch}{1.3}
        \begin{tabular}{lccccc}
            \toprule[1.2pt]
            \multicolumn{6}{c}{\bf Ego4D} \\
            \midrule
            $\mathcal{L}_{neg}$ & $\mathcal{L}_{sup}$ & \bf Perceiver & \bf F1 Score & \bf Recall & \bf Precision \\
            \midrule
            - & - & - & 37.4 & 53.6 & 28.7 \\ \hdashline
            \cmark &  &  & 39.4 & 54.2 & 31.0 \\
             & \cmark &  & 39.4 & 53.5 & 31.2 \\
             &  & \cmark & 39.5 & 53.5 & 31.3 \\
            \midrule
            \rowcolor{Blue}
            \cmark & \cmark & \cmark & \bf 40.1 (+2.7) & \bf 54.1 (+0.5) & \bf 31.9 (+3.2) \\
            \midrule \midrule
            \multicolumn{6}{c}{\bf AEA} \\
            \midrule
            $\mathcal{L}_{neg}$ & $\mathcal{L}_{sup}$ & \bf Perceiver & \bf F1 Score & \bf Recall & \bf Precision \\
            \midrule
            - & - & - & 57.7 & 63.4 & 53.0 \\ \hdashline
            \cmark &  &  & 58.8 & 66.2 & 52.8 \\
             & \cmark &  & 59.4 & 67.0 & 53.4 \\
             &  & \cmark & 59.4 & 64.8 & 54.8 \\
            \midrule
            \rowcolor{Blue}
            \cmark & \cmark & \cmark & \bf 60.3 (+2.6) & \bf 67.2 (+3.8) & \bf 54.7 (+1.7) \\
            \bottomrule[1.2pt]
        \end{tabular}
    }
    \caption{The ablation study to validate the effectiveness of each proposed component.}
    \label{tab3}
\end{table}

%% file: sec/5_disc.tex
\section{Discussion}
\subsection{Computational Cost}
We analyze the computation cost of our framework with respect to the number of parameters, FLOPs, and inference time. The inference time is measured on a single NVIDA L4 GPU by estimating the average processing time per a video input. Table~\ref{tab4} presents the computational cost analysis and comparison with CSTS~\cite{csts}. Since CSTS requires audio sensor data at inference time, it incurs significant computational overhead. In contrast, our method leverages language guidance only during training, resulting in much lower inference-time costs. Therefore, our approach could be more suitable for efficient real-world applications.

\input{tables/table4}

\subsection{Limitation}
A current limitation of our framework is the dependency on scene summary descriptions, which must be prepared before training when no manual annotations are available. These descriptions are generated using pretrained models adding the preprocessing step to the pipeline. A promising future research direction is to integrate this step into an end-to-end framework, minimizing additional computational overhead.

\subsection{Future Work}
Our method can be extended in several promising directions. First, it provides valuable clues for predicting the camera wearer's focus, which can support downstream tasks such as first-person action anticipation, object interaction, and social interaction understanding. Thus, our scene context-aware learning approach could be broadly applied to various egocentric video analysis tasks. Second, our method does not currently utilize other sensory data. When incorporated with complementary modalities such as audio, optical flow, or head movement signals, it could further improve attention prediction performance. In other worlds, it remains a promising avenue for future research to explore such multimodal extensions.

%% file: tables/table4.tex
\definecolor{Blue}{rgb}{0.9, 0.95, 0.97}

\begin{table}[t]
    \centering
    \renewcommand{\tabcolsep}{0.3mm}
    \resizebox{0.99\linewidth}{!}
    {\small
        \renewcommand{\arraystretch}{1.3}
        \begin{tabular}{lccc}
            \toprule[1.2pt]
            \bf Methods & \bf \# Params (M) & \bf FLOPs (G) & \bf Inference time (ms) \\
            \midrule
            CSTS (ECCV'24) & 187.8 & 11.5 & 100.54 \\ \midrule
            \rowcolor{Blue}
            \bf Ours & \bf 85.4 (54.5\%$\downarrow$) & \bf 10.4 (9.6\%$\downarrow$) & \bf 81.98 (18.5\%$\downarrow$) \\
            \bottomrule[1.2pt]
        \end{tabular}
    }
    \caption{The computational cost analysis and comparison with CSTS which adopts audio sensor data at inference time.}
    \label{tab4}
\end{table}

%% file: sec/6_conc.tex
\section{Conclusion}
\label{sec:conc}
In this paper, we introduced a novel language-guided, scene context-aware learning framework for robust egocentric visual attention prediction. To this end, we presented a context perceiver and two training objectives. The context perceiver aims to obtain contextualized video features by capturing the overall scene context of the given video, guided by language-based scene summary descriptions. Based on this, the negative region loss encourages the framework to focus on the target PoI regions, while the region suppression loss reduces distractions from the other attention-irrelevant regions. We conducted extensive experiments and analyses on Ego4D and AEA datasets, and demonstrated its the effectiveness showing state-of-the-art performances on both datasets. We hope that our scene context-aware learning approach can be broadly applied to diverse egocentric vision tasks (e.g., egocentric pose estimation, action recognition, and anticipation), which are essential to understand events and interactions surrounding the first person.

%% file: main.bib
@String(PAMI = {IEEE Trans. Pattern Anal. Mach. Intell.})

@String(CVPR= {IEEE Conf. Comput. Vis. Pattern Recog.})

@String(ICCV= {Int. Conf. Comput. Vis.})

@String(ECCV= {Eur. Conf. Comput. Vis.})

@String(NIPS= {Adv. Neural Inform. Process. Syst.})

@String(BMVC= {Brit. Mach. Vis. Conf.})

@String(TIP  = {IEEE Trans. Image Process.})

@String(ICCVW= {Int. Conf. Comput. Vis. Worksh.})

@String(WACV= {IEEE Winter. Conf. Appl. Comput. Vis.})

@article{lego,
        title={LEGO: Learning EGOcentric Action Frame Generation via Visual Instruction Tuning},
        author={Lai, Bolin and Dai, Xiaoliang and Chen, Lawrence and Pang, Guan and Rehg, James M and Liu, Miao},
        journal={arXiv preprint arXiv:2312.03849},
        year={2023}
      }

@inproceedings{egoexo,
  title={Ego-exo: Transferring visual representations from third-person to first-person videos},
  author={Li, Yanghao and Nagarajan, Tushar and Xiong, Bo and Grauman, Kristen},
  booktitle=CVPR,
  pages={6943--6953},
  year={2021}
}

@article{egovlp,
  title={Egocentric video-language pretraining},
  author={Lin, Kevin Qinghong and Wang, Jinpeng and Soldan, Mattia and Wray, Michael and Yan, Rui and Xu, Eric Z and Gao, Difei and Tu, Rong-Cheng and Zhao, Wenzhe and Kong, Weijie and others},
  journal=NIPS,
  volume={35},
  pages={7575--7586},
  year={2022}
}

@inproceedings{egoexolearn,
  title={EgoExoLearn: A Dataset for Bridging Asynchronous Ego-and Exo-centric View of Procedural Activities in Real World},
  author={Huang, Yifei and Chen, Guo and Xu, Jilan and Zhang, Mingfang and Yang, Lijin and Pei, Baoqi and Zhang, Hongjie and Dong, Lu and Wang, Yali and Wang, Limin and others},
  booktitle=CVPR,
  pages={22072--22086},
  year={2024}
}

@inproceedings{backpack,
  title={A Backpack Full of Skills: Egocentric Video Understanding with Diverse Task Perspectives},
  author={Peirone, Simone Alberto and Pistilli, Francesca and Alliegro, Antonio and Averta, Giuseppe},
  booktitle=CVPR,
  pages={18275--18285},
  year={2024}
}

@article{eye1,
  title={Eye movements in natural behavior},
  author={Hayhoe, Mary and Ballard, Dana},
  journal={Trends in Cognitive Sciences},
  volume={9},
  number={4},
  pages={188--194},
  year={2005},
  publisher={Elsevier}
}

@article{eye2,
  title={Eye movements and their functions in everyday tasks},
  author={Foulsham, Tom},
  journal={Eye},
  volume={29},
  number={2},
  pages={196--199},
  year={2015},
  publisher={Nature Publishing Group}
}

@inproceedings{eye3,
  title={Temporal localization and spatial segmentation of joint attention in multiple first-person videos},
  author={Huang, Yifei and Cai, Minjie and Kera, Hiroshi and Yonetani, Ryo and Higuchi, Keita and Sato, Yoichi},
  booktitle=ICCVW,
  pages={2313--2321},
  year={2017}
}

@article{context1,
  title={Human gaze control during real-world scene perception},
  author={Henderson, John M},
  journal={Trends in Cognitive Sciences},
  volume={7},
  number={11},
  pages={498--504},
  year={2003},
  publisher={Elsevier}
}

@article{context2,
  title={Contextual guidance of eye movements and attention in real-world scenes: the role of global features in object search.},
  author={Torralba, Antonio and Oliva, Aude and Castelhano, Monica S and Henderson, John M},
  journal={Psychological Review},
  volume={113},
  number={4},
  pages={766},
  year={2006},
  publisher={American Psychological Association}
}

@article{context3,
  title={Scene context automatically drives predictions of object transformations},
  author={Aldegheri, Giacomo and Gayet, Surya and Peelen, Marius V},
  journal={Cognition},
  volume={238},
  pages={105521},
  year={2023},
  publisher={Elsevier}
}

@inproceedings{gaze1,
  title={Learning to predict gaze in egocentric video},
  author={Li, Yin and Fathi, Alireza and Rehg, James M},
  booktitle=ICCV,
  pages={3216--3223},
  year={2013}
}

@inproceedings{gaze2,
  title={Digging deeper into egocentric gaze prediction},
  author={Tavakoli, Hamed Rezazadegan and Rahtu, Esa and Kannala, Juho and Borji, Ali},
  booktitle=WACV,
  pages={273--282},
  year={2019},
  organization={IEEE}
}

@article{gaze3,
  title={An ego-vision system for discovering human joint attention},
  author={Huang, Yifei and Cai, Minjie and Sato, Yoichi},
  journal={IEEE Trans. Human Mach. Syst.},
  volume={50},
  number={4},
  pages={306--316},
  year={2020},
  publisher={IEEE}
}

@article{gaze4,
  title={Mutual context network for jointly estimating egocentric gaze and action},
  author={Huang, Yifei and Cai, Minjie and Li, Zhenqiang and Lu, Feng and Sato, Yoichi},
  journal=TIP,
  volume={29},
  pages={7795--7806},
  year={2020},
  publisher={IEEE}
}

@inproceedings{gaze5,
  title={Predicting gaze from egocentric social interaction videos and imu data},
  author={Thakur, Sanket Kumar and Beyan, Cigdem and Morerio, Pietro and Del Bue, Alessio},
  booktitle={Int. Conf. Multimodal Interact.},
  pages={717--722},
  year={2021}
}

@inproceedings{gaze6,
  title={SwinGaze: Egocentric Gaze Estimation with Video Swin Transformer},
  author={Li, Yujie and Wang, Xinghe and Ma, Zihang and Wang, Yifu and Meyer, Michael C},
  booktitle={IEEE Int. Symp. Embedded Multicore/Many-core System-on-Chip (MCSoC)},
  pages={123--127},
  year={2023},
  organization={IEEE}
}

@inproceedings{dfg,
  title={Deep future gaze: Gaze anticipation on egocentric videos using adversarial networks},
  author={Zhang, Mengmi and Teck Ma, Keng and Hwee Lim, Joo and Zhao, Qi and Feng, Jiashi},
  booktitle=CVPR,
  pages={4372--4381},
  year={2017}
}

@article{dfg+,
  title={Anticipating where people will look using adversarial networks},
  author={Zhang, Mengmi and Ma, Keng Teck and Lim, Joo Hwee and Zhao, Qi and Feng, Jiashi},
  journal=PAMI,
  volume={41},
  number={8},
  pages={1783--1796},
  year={2018},
  publisher={IEEE}
}

@inproceedings{attntrans,
  title={Predicting gaze in egocentric video by learning task-dependent attention transition},
  author={Huang, Yifei and Cai, Minjie and Li, Zhenqiang and Sato, Yoichi},
  booktitle=ECCV,
  pages={754--769},
  year={2018}
}

@article{gazemle,
  title={In the eye of the beholder: Gaze and actions in first person video},
  author={Li, Yin and Liu, Miao and Rehg, James M},
  journal=PAMI,
  volume={45},
  number={6},
  pages={6731--6747},
  year={2021},
  publisher={IEEE}
}

@inproceedings{glc,
  title={In the Eye of Transformer: Global-Local Correlation for Egocentric Gaze Estimation},
  author={Lai, Bolin and Liu, Miao and Ryan, Fiona and Rehg, James M},
  booktitle=BMVC,
  year={2022}
}

@inproceedings{csts,
  title={Listen to look into the future: Audio-visual egocentric gaze anticipation},
  author={Lai, Bolin and Ryan, Fiona and Jia, Wenqi and Liu, Miao and Rehg, James M},
  booktitle=ECCV,
  year={2024}
}

@inproceedings{mvit,
  title={Multiscale vision transformers},
  author={Fan, Haoqi and Xiong, Bo and Mangalam, Karttikeya and Li, Yanghao and Yan, Zhicheng and Malik, Jitendra and Feichtenhofer, Christoph},
  booktitle=ICCV,
  pages={6824--6835},
  year={2021}
}

@inproceedings{ego4d,
  title={Ego4d: Around the world in 3,000 hours of egocentric video},
  author={Grauman, Kristen and Westbury, Andrew and Byrne, Eugene and Chavis, Zachary and Furnari, Antonino and Girdhar, Rohit and Hamburger, Jackson and Jiang, Hao and Liu, Miao and Liu, Xingyu and others},
  booktitle=CVPR,
  pages={18995--19012},
  year={2022}
}

@article{aria,
  title={Aria Everyday Activities Dataset},
  author={Lv, Zhaoyang and Charron, Nickolas and Moulon, Pierre and Gamino, Alexander and Peng, Cheng and Sweeney, Chris and Miller, Edward and Tang, Huixuan and Meissner, Jeff and Dong, Jing and others},
  journal={arXiv preprint arXiv:2402.13349},
  year={2024}
}

@inproceedings{egovis1,
  title={Egocentric scene understanding via multimodal spatial rectifier},
  author={Do, Tien and Vuong, Khiem and Park, Hyun Soo},
  booktitle=CVPR,
  pages={2832--2841},
  year={2022}
}

@misc{egovis2,
      title={Eliciting In-Context Learning in Vision-Language Models for Videos Through Curated Data Distributional Properties},
      author={Keunwoo Peter Yu and Zheyuan Zhang and Fengyuan Hu and Shane Storks and Joyce Chai},
      year={2024},
      eprint={2311.17041},
      archivePrefix={arXiv},
      primaryClass={cs.CV},
      url={https://arxiv.org/abs/2311.17041},
}

@inproceedings{egovis3,
  title={Assembly101: A large-scale multi-view video dataset for understanding procedural activities},
  author={Sener, Fadime and Chatterjee, Dibyadip and Shelepov, Daniel and He, Kun and Singhania, Dipika and Wang, Robert and Yao, Angela},
  booktitle=CVPR,
  pages={21096--21106},
  year={2022}
}

@article{epic,
  title={The epic-kitchens dataset: Collection, challenges and baselines},
  author={Damen, Dima and Doughty, Hazel and Farinella, Giovanni Maria and Fidler, Sanja and Furnari, Antonino and Kazakos, Evangelos and Moltisanti, Davide and Munro, Jonathan and Perrett, Toby and Price, Will and others},
  journal=PAMI,
  volume={43},
  number={11},
  pages={4125--4141},
  year={2020},
  publisher={IEEE}
}

@inproceedings{egovis4,
  title={Scene-aware egocentric 3d human pose estimation},
  author={Wang, Jian and Luvizon, Diogo and Xu, Weipeng and Liu, Lingjie and Sarkar, Kripasindhu and Theobalt, Christian},
  booktitle=CVPR,
  pages={13031--13040},
  year={2023}
}

@inproceedings{egovis5,
  title={Leveraging next-active objects for context-aware anticipation in egocentric videos},
  author={Thakur, Sanket and Beyan, Cigdem and Morerio, Pietro and Murino, Vittorio and Del Bue, Alessio},
  booktitle=WACV,
  pages={8657--8666},
  year={2024}
}

@article{egovis6,
  title={Egocentric Scene-aware Human Trajectory Prediction},
  author={Wang, Weizhuo and Liu, C Karen and Kennedy III, Monroe},
  journal={arXiv preprint arXiv:2403.19026},
  year={2024}
}

@inproceedings{egovis7,
  title={Scan context: Egocentric spatial descriptor for place recognition within 3d point cloud map},
  author={Kim, Giseop and Kim, Ayoung},
  booktitle={IEEE/RSJ Int. Conf. Intell. Robots. Syst.},
  pages={4802--4809},
  year={2018},
  organization={IEEE}
}

@inproceedings{egovis8,
  title={With a Little Help from my Temporal Context: Multimodal Egocentric Action Recognition},
  author={Kazakos, Evangelos and Huh, Jaesung and Nagrani, Arsha and Zisserman, Andrew and Damen, Dima},
  booktitle=BMVC,
  year={2021}
}

@article{egovis9,
  title={Shaping embodied agent behavior with activity-context priors from egocentric video},
  author={Nagarajan, Tushar and Grauman, Kristen},
  journal=NIPS,
  volume={34},
  pages={29794--29805},
  year={2021}
}

@article{egovis10,
  title={EgoEnv: Human-centric environment representations from egocentric video},
  author={Nagarajan, Tushar and Ramakrishnan, Santhosh Kumar and Desai, Ruta and Hillis, James and Grauman, Kristen},
  journal=NIPS,
  volume={36},
  year={2024}
}

@article{videochat2,
  title={Videochat: Chat-centric video understanding},
  author={Li, KunChang and He, Yinan and Wang, Yi and Li, Yizhuo and Wang, Wenhai and Luo, Ping and Wang, Yali and Wang, Limin and Qiao, Yu},
  journal={Science China Information Sciences},
  volume={68},
  number={10},
  pages={200102},
  year={2025},
  publisher={Springer}
}

@article{nv-embed-v2,
  title={NV-Embed: Improved Techniques for Training LLMs as Generalist Embedding Models},
  author={Lee, Chankyu and Roy, Rajarshi and Xu, Mengyao and Raiman, Jonathan and Shoeybi, Mohammad and Catanzaro, Bryan and Ping, Wei},
  journal={arXiv preprint arXiv:2405.17428},
  year={2024}
}

@misc{slowfast,
  author =       {Haoqi Fan and Yanghao Li and Bo Xiong and Wan-Yen Lo and
                  Christoph Feichtenhofer},
  title =        {PySlowFast},
  howpublished = {\url{https://github.com/facebookresearch/slowfast}},
  year =         {2020}
}

@inproceedings{i3d,
  title={Slowfast networks for video recognition},
  author={Feichtenhofer, Christoph and Fan, Haoqi and Malik, Jitendra and He, Kaiming},
  booktitle={Proceedings of the IEEE/CVF international conference on computer vision},
  pages={6202--6211},
  year={2019}
}
